\documentclass[11pt,journal]{IEEEtran}

\usepackage{times}
\usepackage{epsfig}
\usepackage{graphicx}
\usepackage{amsmath}
\usepackage{amssymb}
\usepackage{hyperref}

\begin{document}

\title{TriResNet: A Deep Triple-stream Residual Network for Histopathology Grading}

\author{Rene Bidart and Alexander Wong\\
Waterloo Artificial Intelligence Insitute\\
Department of Systems Design Engineering\\
University of Waterloo\\
Waterloo, Ontario\\
{\tt\small \{rbbidart,a28wong\}@uwaterloo.ca}
}

\maketitle

\begin{abstract}
While microscopic analysis of histopathological slides is generally considered as the gold standard method for performing cancer diagnosis and grading, the current method for analysis is extremely time consuming and labour intensive as it requires pathologists to visually inspect tissue samples in a detailed fashion for the presence of cancer.  As such, there has been significant recent interest in computer aided diagnosis systems for analysing histopathological slides for cancer grading to aid pathologists to perform cancer diagnosis and grading in a more efficient, accurate, and consistent manner.  In this work, we investigate and explore a deep triple-stream residual network (\textbf{TriResNet}) architecture for the purpose of tile-level histopathology grading, which is the critical first step to computer-aided whole-slide histopathology grading.  In particular, the design mentality behind the proposed TriResNet network architecture is to facilitate for the learning of a more diverse set of quantitative features to better characterize the complex tissue characteristics found in histopathology samples.  Experimental results on two widely-used computer-aided histopathology benchmark datasets (CAMELYON16 dataset and Invasive Ductal Carcinoma (IDC) dataset) demonstrated that the proposed TriResNet network architecture was able to achieve noticeably improved accuracies when compared with two other state-of-the-art deep convolutional neural network architectures. Based on these promising results, the hope is that the proposed TriResNet network architecture could become a useful tool to aiding pathologists increase the consistency, speed, and accuracy of the histopathology grading process.
\end{abstract}


\begin{figure}[t]
\begin{center}
   \includegraphics[width=1\linewidth]{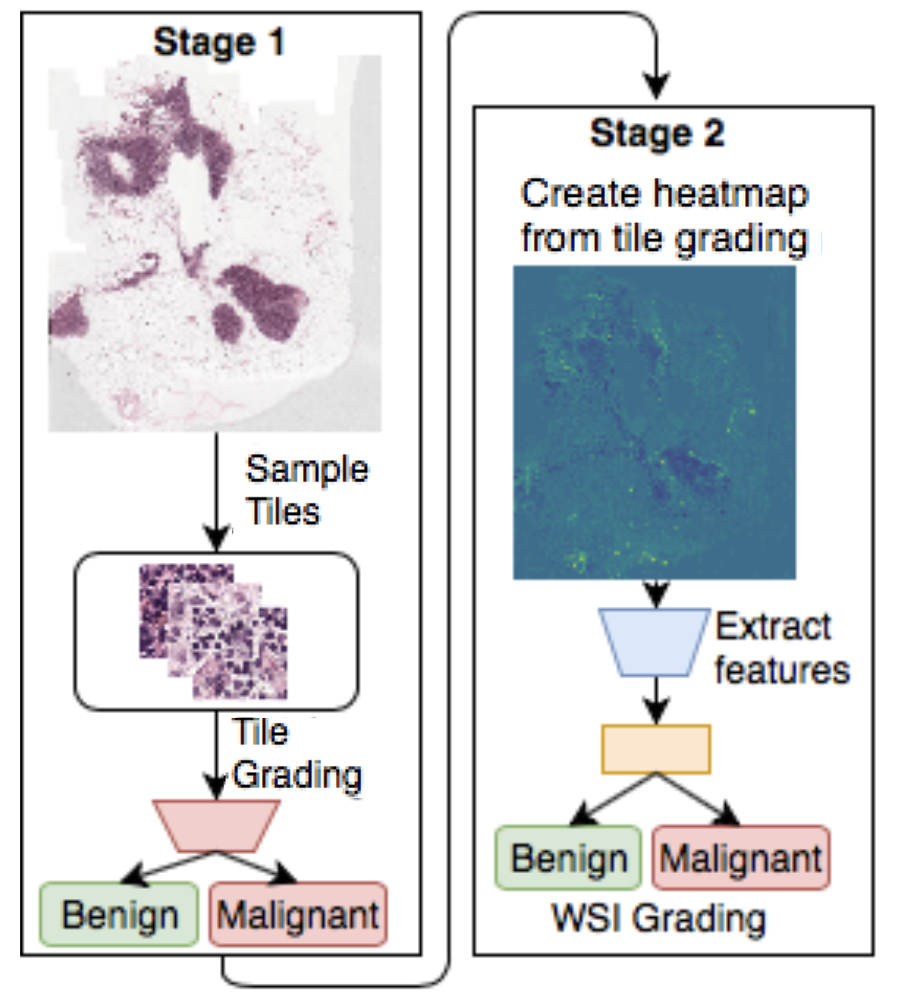}
\end{center}
   \caption{\label{fig:wsi_full_arc}\textit{An overview of the deep learning-driven computer-aided whole slide image (WSI) histopathology grading pipeline. In the first stage, tiles from the WSI are extracted and tile-level histopathology grading is performed using a deep convolutional neural network (CNN). In the second stage, the histopathology gradings for all tiles in the WSI are combined to create a malignancy probability heatmap.  Features are then extracted from this heatmap, and are used to generate a final WSI-level grading. In this paper, we focus on improving tile-level histopathology grading using the proposed TriResNet network architecture.}}
\end{figure}

\section{Introduction}

The microscopic analysis of hematoxylin and eosin (H\&E) stained histopathological slides is generally considered as the gold standard method for diagnosing and grading cancers \cite{kumar2014robbins}\cite{fischer2008hematoxylin}.  However, the current method for performing such an analysis requires the manual visual inspection of human pathologists, and as such can be limiting in several aspects.  First of all, histopathological diagnosis and grading via manual visual inspection relies on the qualitative analysis of images from a microscope by a human pathologist, and as such can suffer from high inter-observer and intra-observer variability, particularly with the lack of standardization in the diagnosis and grading process.  Second, the visual inspection of histopathological slides is an extremely time-consuming and labor intensive process, especially considering the large volume of slides that a typical pathologist must analyze, with each slide containing millions of cells~\cite{allende2014inter,raab2005clinical,elmore2015diagnostic}.  These issues associated with the current method for performing histopathological diagnosis and grading are problematic in developed countries, but are far greater in developing countries where there is a severe lack of trained pathologists\cite{benediktsson2007pathology}.  As such, there has been significant recent interest in computer aided diagnosis systems for analysing histopathological slides for cancer grading to aid pathologists to perform cancer diagnosis and grading in a more efficient, accurate, and consistent manner.

Amongst the different strategies proposed for the purpose of computer aided histopathology grading, one of the most promising recent developments has been the leveraging of machine learning for building computational predictive models learnt directly from the wealth of histopathological slides.  Earlier methods that leveraged machine learning for computer aided histopathology grading utilized human-engineered quantitative features extracted from a histopathological image, followed by the application of a machine learning-driven classification model on these extracted features~\cite{monaco2010high,he2012histology,altunbay2010color,petushi2006large,doyle2008automated,basavanhally2013multi}. For example,~\cite{petushi2006large} utilized tissue texture features by performing cell segmentation and calculating nuclei density and position as extracted features to be fed into a quadratic classifier. Other approaches have utilized a heterogeneous mix of human-engineered features ranging from simple features such as hue, saturation, and intensity to more complex features such as texture-based features (e.g., Haralick features and Gabor features) as well as graph-based features, followed by the application of support vector machines on these extracted features~\cite{doyle2008automated}.  However, such earlier methods that leverage human-engineered features have been limited in their performance due to the significant difficulties for human experts to manually design a comprehensive set of features that can comprehensively capture the complex tissue characteristics exhibited in histopathological slides. Therefore, methods that can learn a comprehensive set of important quantitative features for discriminating between benign and cancerous tissue directly from histopathological slides themselves is highly desired.

In recent years, the concept of deep learning~\cite{lecun2015deep} has revolutionized the area of computer-aided histopathology diagnosis and grading by automatically learning discriminative quantitative features from the wealth of available histopathological images in a direct manner, rather than being constrained by the limitations of human-engineered features.  In particular, a type of deep neural network known as deep convolutional neural networks (CNN), which demonstrated state-of-the-art performance on visual perception compared to other machine learning algorithms~\cite{alexnet}, has been leveraged for computer-aided histopathology grading to great success~\cite{wang2016deep,shah2017deep,liu2017detecting,Li2018,Lunit}. These deep learning-driven computer-aided whole slide image (WSI) histopathology grading approaches tend to share the same general pipeline.  More specifically, the vast size of whole slide image make them computational intractable to be processed by a deep convolutional neural network in a single inference pass, as is commonly performed in general image classification where the images are significantly smaller in size.

To handle this size and complexity issue, these approaches breaks the task of WSI histopathology grading into two main stages (an overview of this two-stage approach is illustrated in Fig.~\ref{fig:wsi_full_arc}).  In the first stage, tissue image tiles are extracted from the WSI after preprocessing is used to reduce the irrelevant white space in the slide. A CNN trained to perform tile-level tissue grading is then used to grade each of the individual tiles extracted over the entire WSI. In the second stage, the histopathology gradings for all tiles in the WSI are then combined together to create a malignancy probability heatmap, and from this heatmap a number of WSI-level geometrical and morphological features are then extracted. These extracted WSI-level features are then used by a machine learning classification model to generate the final WSI-level histopathology grading.  As such, improvements to either of these two stages would yield benefits for the overall WSI histopathology grading processing.  In this paper, we place our focus on improving the first stage of the computer-aided WSI histopathology grading pipeline by improving the performance of the tile-level tissue grading process through the introduction of an improved CNN network architecture.

The key contribution of this paper is the introduction of a novel deep triple-stream residual network (TriResNet) architecture  for the purpose of improved tile-level histopathology grading.  The proposed TriResNet architecture incorporates three different streams comprised of a deep stack of residual blocks, with the underlying motivation that, through careful training, each residual stream will learn a different set of quantitative features for better characterizing different aspects of the complex tissue characteristics captured in histopathology images than what can be achieved by a single-stream network.  A multi-stage targeted training procedure is also introduced to overcome the difficulty of training such a large network architecture as well as better encouraging feature diversity within the network.

This paper is organized as follows.  Section 2 provides a detailed description of the proposed TriResNet network architecture [2.2], as well as the multi-stage targeted training policy leveraged to train the network [2.3]. Section 3 presents the experiments conducted on two publicly available histopathology benchmark datasets (CAMELYON16 dataset and Invasive Ductal Carcinoma (ICD) dataset), including a description of datasets [3.1] as well as the experimental setup [3.2].  Finally, experimental results and discussion of the results are presented in Section 4, with conclusions drawn and future work discussed in Section 5.

\section{Deep Triple-stream Residual Network}

The proposed deep triple-stream residual network (TriResNet) architecture is designed based on the idea of extracting a more diverse set of discriminative quantitative features for better characterizing the diverse and complex tissue characteristics exhibited in histopathological images.  A more detailed description of the proposed network architecture as well as the multi-stage targeted training policy used to train this network is provided below.

\begin{figure}[t]
\begin{center}
   \includegraphics[width=1\linewidth]{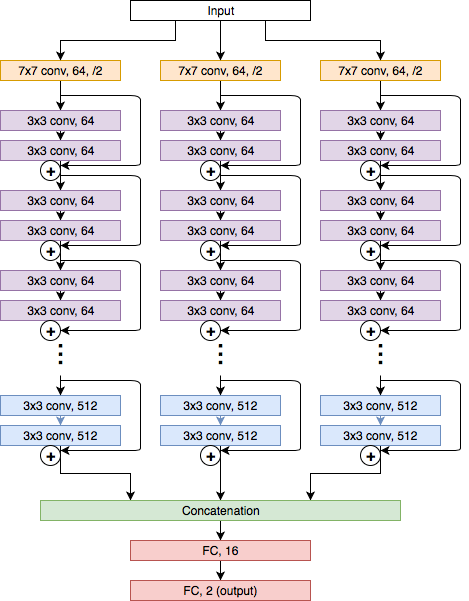}
\end{center}
   \caption{\label{fig:TriResNet}\textit{Deep triple-stream residual network (TriResNet) architecture. An input layer feeds three separate residual streams, which each residual stream composed of a deep stack of residual blocks with a total of 34 layers. Each of these residual streams extract different sets of quantitative features, which are then fed into a concatenation layer, followed by two fully connected layers that result in a final grading prediction of the input histopathological image tile as being either malignant or benign tissue.}}
\end{figure}

\subsection{Network Architecture}
The underlying goal behind the design of the proposed TriResNet network architecture is to better learn a larger, more diverse set of quantitative features for characterizing complex and varied tissue characteristics exhibited in histopathology images.  To achieve this goal, we leverage the notion of residual learning first proposed in~\cite{he2016deep}, which has not only demonstrated state-of-the-art performance for a wide variety of applications such as general image recognition \cite{he2016deep}, but has been recognized in research literature for its terrific ability to perform well both for feature extraction and fine-tuning\cite{kornblith2018better}.

The strategy we leverage in the proposed TriResNet network to encourage greater feature diversity is to incorporate multiple streams of residual blocks, with the underlying premise being that each of these independent streams, when trained appropriately, will be able to capture different nuances within the tissue characteristics in histopathology images.  More specifically, as shown in Fig.~\ref{fig:TriResNet}, the proposed TriResNet architecture comprises of three separate residual streams consisting of deep stacks of residual blocks for a total of 34 layers in each stream, with each stream made up of a different set of learned weights to capture diverse feature sets. To explicitly encourage feature diversity of individual residual streams and learn to capture different tissue nuances amongst the streams, each residual stream undergoes pre-training exposure to different data collections, which will be described in detail in Section [2.3] where the multi-stage targeted training policy is outlined.

The features of the last residual blocks in each of the three residual streams within the proposed TriResNet architecture are combined in a concatenation layer, which is then fed into a 16-neuron fully connected layer, followed by a ReLU layer, which then feeds into a final fully connected layer where the number of neurons is equal to the number tissue grades to provide the final prediction output.

\subsection{Multi-stage Targeted Training Policy}
One of the challenges with leveraging the proposed TriResNet architecture for tile-level histopathology grading is that training such a complex network is extremely difficult because of the large number of parameters within this network (which makes converging to an appropriate solution quite challenging given this large parameter size) as well as the dangers of over-fitting. In order to tackle the aforementioned problem, we leverage a multi-stage targeted training policy consisting of the following three main stages:
\begin{enumerate}
  \item Targeted pre-training of individual residual streams.
  \item Targeted training of fully connected network layers.
  \item End-to-end fine-tuning of the full network.
\end{enumerate}
~\\
\noindent\textbf{Targeted pre-training of individual residual streams}. As the first stage of the training policy, each of the three residual streams are pre-trained independently by freezing the weights of the other residual streams, and augmenting a proxy fully connected output layer to the particular residual stream we wish to pre-train. This ensures that only one particular residual stream will be pre-trained at a time, leaving the other residual streams unaffected during the individual pre-training processes.  Given our goal is to encourage diverse feature learning to better model the diverse tissue characteristics in histopathology, we utilize a stochastic pre-training policy for each of the residual streams via their individual proxy fully connected output layers such that the individual streams are exposed to different random batches of tissue tiles during the pre-training process.  This ensures that each residual stream will converge to a different set of weights, and thus will be capable of capturing a diverse set of quantitative features compared to the other residual streams during inference. To speed up the pre-training process, each residual stream was initialized with pre-trained weights based on the ImageNet Challenge Dataset~\cite{imagenet_cvpr09}.  This pre-training process is repeated for each of the residual streams, and the proxy layers are removed at the end of the pre-training processes. \\

\noindent\textbf{Targeted training of fully connected network layers}. After the targeted training of individual residual streams, we now focus on the targeted training of the fully connected network layers. The rationale behind this is that, because the random initialization of these layers, there is strong potential for convergence issues if the entire network architecture is trained end-to-end at this point.  By freezing the weights of the individual residual streams while the fully connected layers begin to learn, we allow the fully connected layers to converge to a good set of weights without the convergence issues associated with training the entire network at this point.  Furthermore, because backpropagation is performed only on the fully connected layers, the time to convergence is greatly accelerated.\\

\noindent\textbf{End-to-end fine-tuning of the full network}. After the fully-connected layers in the network have been trained using the targeted training process, the entire network undergoes an end-to-end fine-tuning process to further improve the performance of the full network architecture. In this part of the training process, we backpropogate the gradient though the entire network, including each of the three residual streams and the fully connected layers as a whole, thus optimizing the weights of the entire network. This is done at a lower learning rate that is a factor of 10 times lower than the initial learning rate.  This end-to-end fine-tuning phase also encourages the individual residual streams to work cohesively together as a complete network architecture.\\

\section{Experiments}
To study the efficacy of the proposed TriResNet network for the purpose of tile-level histopathology grading, we performed a series of experiments using two widely-used histopathological image benchmark datasets.  The details of these datasets as well as the experimental setup are presented below.

\subsection{Data}
We investigate two publicly-available histopathological image benchmark datasets: i) CAMELYON16\cite{camelyon} dataset, and ii) Invasive Ductal Carcinoma (IDC)~\cite{janowczyk2016deep,cruz2014automatic} dataset.

\subsubsection{CAMELYON16 Dataset}
\begin{figure}[t]
\begin{center}
   \includegraphics[width=1\linewidth]{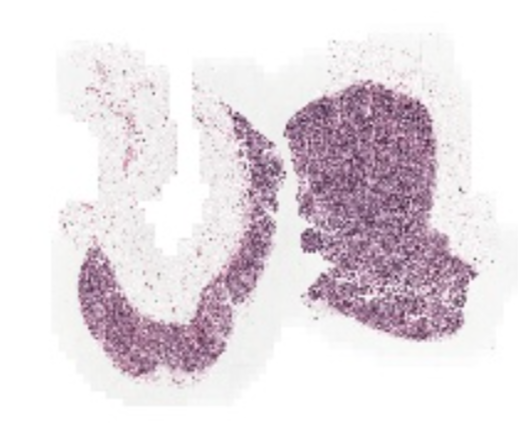}
\end{center}
   \caption{\label{fig:wsi_crop}\textit{An example of a whole slide image from the CAMELYON16 dataset, showing the large amount of irrelevant background in white.}}
\end{figure}

The CAMELYON16 dataset contains lymph node tissues of breast cancer patients, with the goal being to find metastasis of breast cancer. CAMELYON16 consists of 400 whole slide images (WSI) divided into 270 for training and 160 for testing. Ground truth is provided by a mask corresponding to each slide, which is an image with pixel level annotation indicating the cancerous regions. Both the mask and the WSI are very high resolution (100,000 x 200,000 pixels), with a single file being about 5 gigabytes. These are stored in a multi-resolution format, meaning that each file contains the high resolution image, as well as down sampled versions to a minimum size of about 512x1024. An example WSI from the CAMELYON16 dataset is shown in Fig.~\ref{fig:wsi_crop}.

Due to the large size of the high resolution WSI slides making it difficult to handle and even perform simple operations on the slides in a direct manner, OpenSlide\cite{openslide} is used to read in subsections of the WSI at a lower resolution.

\subsubsection{Preprocessing}
Because of the large size of the WSIs, the background is segmented from the actual tissue to greatly reduce the computational requirements in dealing with the histopathological images.  This is accomplished in this paper using the preprocessing approach described in~\cite{wang2016deep}:
\begin{enumerate}
  \item Read in WSI at resolution about 3072x7168 pixels, and convert from RGB to HSV.
  \item Use Otsu's algorithm~\cite{4310076} to separate the background from foreground, then take the union of the result with the H and S channels to generate a tissue mask.
\end{enumerate}

\subsubsection{Dataset Generation}
\begin{figure}[t]
\begin{center}
   \includegraphics[width=1\linewidth]{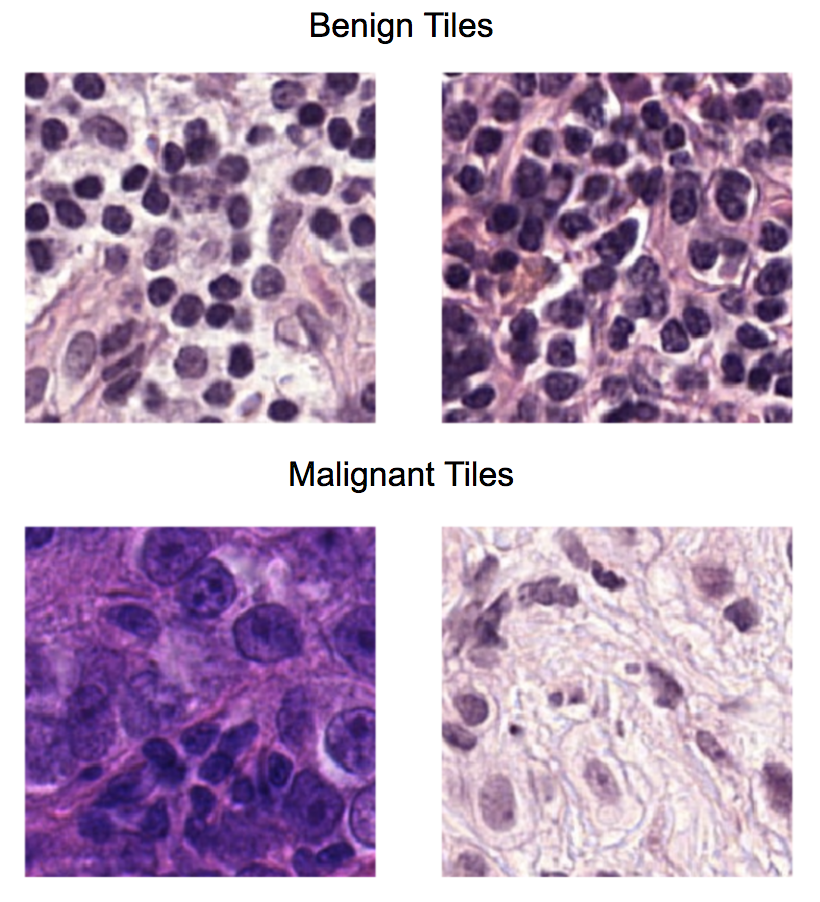}
\end{center}
   \caption{\label{fig:benign-mal-cam}\textit{Examples of tissue image tiles from the CAMELYON16 dataset showing benign and malignant tissues. These tissue image tiles are extracted at the highest magnification (40x), with a size of 224x224 pixels.}}
\end{figure}

In many circumstances, 270 images would be  considered too few data points to train a CNN.  However, due to the fact that we have pixel-level annotations and very high resolution images, a much larger training dataset can be generated from subsections of the original WSI slides and the labels from the pixel-level annotations, using a similar approach as in previous research literature~\cite{dlpath,wang2016deep}.  More specifically, we create a dataset of 224x224 sized tissue image tiles at the highest magnification available, as past research literature has shown that it is most useful to look at the WSIs at the highest magnification~\cite{wang2016deep,liu2017detecting}. Example tissue image tiles obtained from the CAMELYON16 dataset are shown in Fig.~\ref{fig:benign-mal-cam}.

\subsubsection{Class Imbalance}
Given that the WSIs contain far more benign than malignant tissue, this can lead to a significant data imbalance problem when training CNNs.  Therefore, we oversample the malignant class to create a balanced dataset of half malignant and half benign tiles.

To generate the dataset, we alternate sampling between malignant and benign tissue. In the case of malignant tissue, we select a malignant slide and sample from the region indicated by the malignancy mask. In the case of benign tissue, because malignant slides also contain benign tissue, we select any slide, and make sure that the area we sample in is inside the tissue mask and but not in the malignancy mask.

In both cases the malignancy mask is down-sampled to be the same magnification as the tissue mask, and sampling is done at this magnification. These points are then converted to the highest magnification, and we read in the tile at this level. No color normalization is used because it proved to be ineffective in other research~\cite{liu2017detecting}.

\subsubsection{Invasive Ductal Carcinoma (IDC) Dataset}
\begin{figure}[t]
\begin{center}
   \includegraphics[width=1\linewidth]{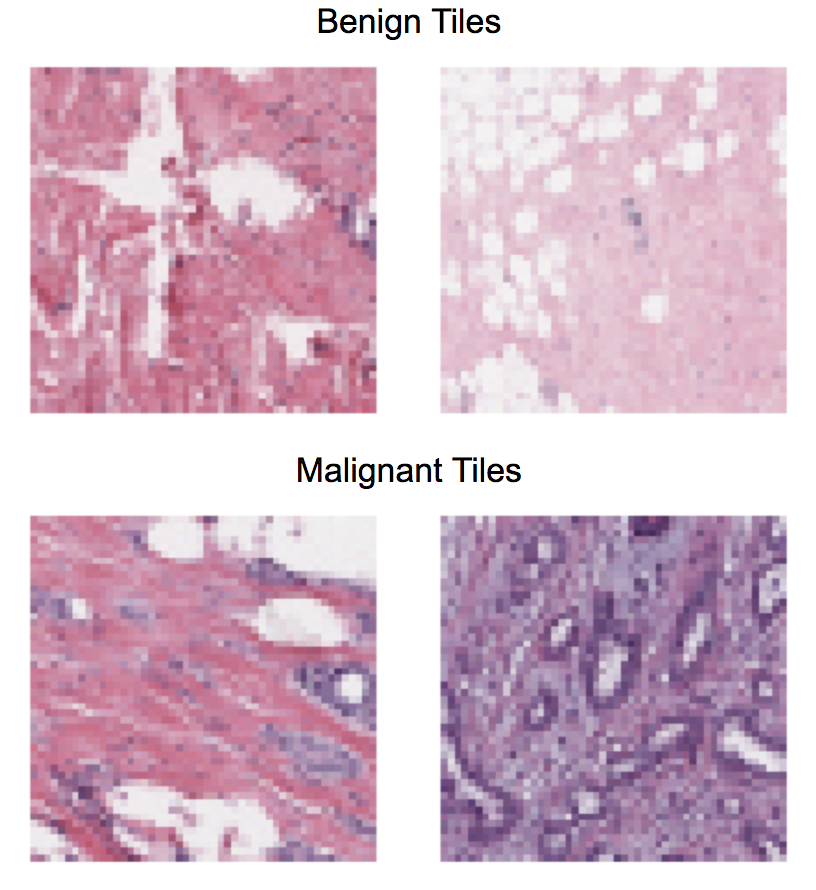}
\end{center}
   \caption{\label{fig:idc_example}\textit{Examples of tissue image tiles from the Invasive Ductal Carcinoma (IDC) dataset showing  benign and malignant tissues. These tissue image tiles are extracted at 40x magnification, with a size of 50x50 pixels.}}
\end{figure}

The Invasive Ductal Carcinoma (IDC) dataset~\cite{janowczyk2016deep,cruz2014automatic} is generated from 162 whole slide images of breast cancer, scanned at 40x magnification. From these slides, 198,738 images were sampled of size 50x50 pixels, with 78,786 of these images containing IDC. Because of the small size of these images, they were resized to the minimum acceptable size for the network during training (197x197), as is done in previous literature. The dataset is already in the format of a standard image classification dataset, so no special preprocessing is needed. Example tissue image tiles from the IDC dataset are shown in figure~\ref{fig:idc_example}.

\subsection{Experimental Setup}
In this paper, we compare the proposed TriResNet network to two state-of-the-art deep convolutional neural networks in order to gauge its performance for the purpose of tile-level histopathology grading. Details of these experiments are shown below.\\

\noindent\textbf{Test-Train-Validation Split} \\
The CAMELYON16 dataset was split between by WSI into 80\% for training and 20\% for validation, and the independent test set given by the CAMELYON16 competition was used for testing and evaluation. For each of the WSIs we extracted malignant and benign tiles, balancing the number of benign and malignant samples, as described above.  For the IDC dataset, a split of 70\% training, 15\% for validation, and 15\% for test set was used.\\

\noindent\textbf{Tested Deep Convolutional Neural Networks} \\
To evaluate and compare the performance of the proposed TriResNet, we compare it to two state-of-the-art deep convolutional neural networks: i) Inception-v3~\cite{szegedy2016rethinking}, and ii) ResNet-34~\cite{he2016deep}. We optimized the performance for the two tested networks to the best of our abilities for high histopathology grading performance on the two datasets.\\

\begin{itemize}
\item \textbf{Inception-v3}: This state-of-the-art network architecture was shown to achieve state-of-the-art performance on the CAMELYON16 histopathology competition~\cite{liu2017detecting}. Because of this network's demonstrated impressive performance in computer-aided histopathology diagnosis, this network is a good choice for comparing the overall performance of the proposed TriResNet network to.\\

\item \textbf{ResNet-34}: This state-of-the-art network architecture is compared with the proposed TriResNet as it was shown to provide state-of-the-art performance on histopathology grading~\cite{shah2017deep,Li2018}.  Furthermore, it was compared with the proposed TriResNet also to get a clearer idea of the benefits of a triple-stream network architecture in capture more diverse features for improved performance when compared to a single-stream network architecture.  The ResNet-34 network has the same number of residual blocks and layers as a single residual stream of TriResNet, thus making the comparison more direct in terms of potential benefits.\\
\end{itemize}

All tested networks are implemented using PyTorch\cite{paszke2017automatic}, and  were initialized with pre-trained weights on the ImageNet Challenge Dataset~\cite{imagenet_cvpr09} to improve the speed of convergence. The Adam optimizer~\cite{kingma2014adam} was used for training. Data augmentation was relatively standard, with random flips, rotations, and brightness transformations.\\

\noindent\textbf{Performance Metrics} \\
For both the CAMELYON16 and IDC datasets, we evaluated the performance of each tested network on their ability to grade tissue image tiles as either malignant or benign. For each network and dataset we evaluated the following three performance metrics on the test set:
\begin{enumerate}
  \item $sensitivity=TP/(TP + FN)$
  \item $specificity=TN/(TN + FP)$
  \item $accuracy=(TP+TN)/(TP + TN + FP + FN)$
\end{enumerate}

\section{Results and Discussion}

Table~\ref{tab:CAMELYON_perf} and Table~\ref{tab:IDC_perf} show the grading performance (in terms of accuracy, sensitivity, and specificity) of the tested networks for the test sets of the CAMELYON16 dataset and the IDC dataset, respectively.  A number of interesting observations can be made from the quantitative results. First of all, it can be clearly seen that for both benchmark datasets, the proposed TriResNet network achieved improved overall accuracy compared to both the tested Inception-v3 and ResNet-34 networks.

When compared to the Inception-v3 network, the proposed TriResNet achieves higher overall accuracy on both datasets, with an increase of 1.8\% and 1.7\% for the CAMELYON16 dataset and the IDC dataset, respectively.  Since this network has been demonstrated to provide strong performance in histopathology image grading~\cite{liu2017detecting} as well as general image classification problems\cite{szegedy2016rethinking}, this demonstrates that the proposed TriResNet can be a very effective network for histopathology grading.

The comparison with the ResNet-34 network demonstrates that a multi-stream network architecture clearly has merits compared to a single-stream network architecture for histopathology grading in terms of capturing a more diverse and discriminative set of quantitative features for characterizing tissue complexities in histopathology images, with the accuracy of the proposed TriResNet being higher by 3.6\% and 1.2\% on the CAMELYON16 and IDC datasets, respectively.

It can also be observed that the proposed TriResNet achieved higher sensitivity and specificity compared to the other tested networks for the IDC dataset, which illustrates the efficacy of the proposed network.  Furthermore, what is particularly interesting to note is, while achieving lower specificity compared to the other tested networks, the increase in sensitivity achieved by the proposed TriResNet is quite pronounced for the CAMELYON16 dataset, where the sensitivity achieved by the proposed TriResNet network is 6.3\% and 9.1\% compared to Inception-v3 and ResNet-34, respectively.  The higher sensitivity achieved by the proposed TriResNet network is particularly important for the case of histopathology grading, as it is more important to identify all instances of malignancy than to have a very low number of false positives, because of the risks associated with missed malignant tissues leading to patients not being treated for malignant cancer.

\begin{figure}[t]
\begin{center}
   \includegraphics[width=1\linewidth]{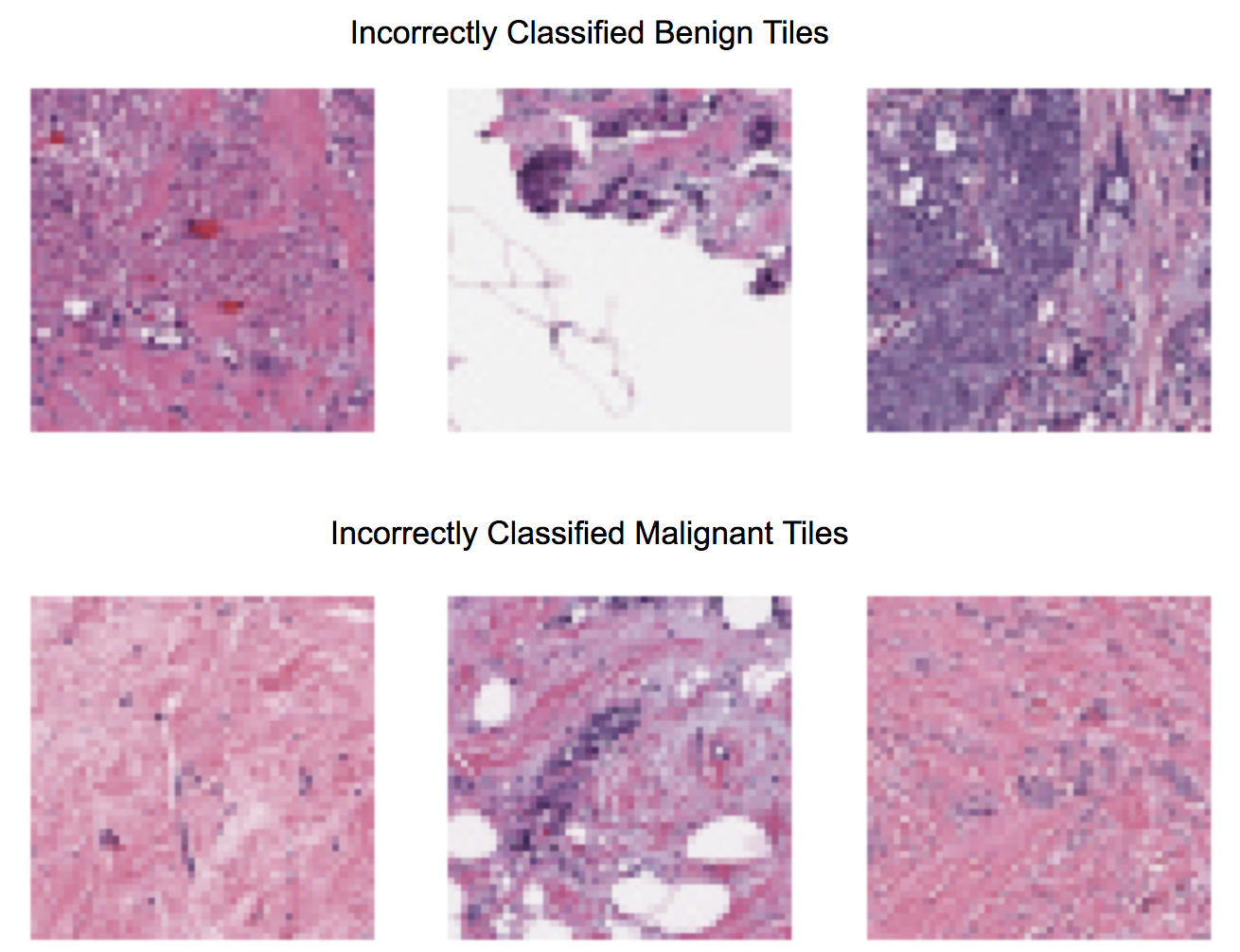}
\end{center}
   \caption{\label{fig:incorrect-idc}\textit{Examples of tissue image tiles that were misclassified by both TriResNet and ResNet-34. The top row shows examples that are actually benign tissue, but were falsely considered malignant, and the bottom set shows malignant tissue that was considered benign.}}
\end{figure}

Rather than simply discuss the strengths of the proposed TriResNet network for the purpose of tile-level histopathology grading, we also study the limitations of its abilities by looking at some example tissue image tiles that are incorrectly graded by the proposed TriResNet, as shown in Fig.~\ref{fig:incorrect-idc}.  It can be observed that both the proposed TriResNet as well as the ResNet-34 network have some systematic difficulties grading in certain circumstances.  For example, one repeated issue experienced by both TriResNet and ResNet-34 was the difficulty associated with grading when there was a large amount of adipose tissue. In addition to this, the networks experienced difficulties when the color of the tissue is different from what is considered the norm; for example, as more malignant tissues tend to look more purple, the networks falsely used this as an indication of malignancy in some benign tissues. Greater diversity of tissues and stains used during training of the networks should alleviate these issues.  Finally, it is important to note that while the proposed TriResNet network achieves very strong performance compared to the other tested networks, it is also noticeably larger in terms of network size compared to the other networks, although for clinical purposes accuracy is in general more important than inference speed.

\begin{table}
\begin{center}
  \caption{Comparison of tested networks on tile-level grading for the test set of the CAMELYON16 dataset. Numbers shown indicate test set performance, and best performance for each category is highlighted in \textbf{bold}}
  \label{tab:CAMELYON_perf}
    \begin{tabular}{ p{2cm} p{1.5cm} p{1.5cm} p{1.5cm}}
    \hline
    \textbf{Network} & \textbf{Accuracy} & \textbf{Sensitivity} & \textbf{Specificity}\\ \hline
    Inception-v3   & 85.3\% & 75.9\% & 95.9\% \\
    ResNet-34       & 83.5\%  & 73.1\% & \textbf{96.5}\% \\
    TriResNet  & \textbf{87.1}\% & \textbf{82.2}\% & 91.2\% \\
    \hline
    \end{tabular}
  \end{center}
  \end{table}

\begin{table}
\begin{center}
  \caption{Comparison of tested networks on tile-level grading for the test set of the IDC dataset. Numbers shown indicate test set performance, and best performance for each category is highlighted in \textbf{bold}}
  \label{tab:IDC_perf}
    \begin{tabular}{ p{2cm} p{1.5cm} p{1.5cm} p{1.5cm}}
    \hline
    \textbf{Network} & \textbf{Accuracy} & \textbf{Sensitivity} & \textbf{Specificity}\\ \hline
    Inception-v3   & 89.2\% & 91.4\% &  83.1\% \\
    ResNet-34       & 89.7\% & 92.3\% & 82.9\% \\
    TriResNet  & \textbf{90.9}\% & \textbf{93.1}\%& \textbf{85.1}\% \\
    \hline
    \end{tabular}
  \end{center}
  \end{table}

\section{Conclusion and Future Work}
In this paper, we introduced a deep triple-stream residual network (TriResNet) architecture designed to better learn more diverse and discriminative features for characterizing complex tissue characteristics, and thus provide improved tile-level histopathology grading.  Experimental results across two widely-used benchmark datasets demonstrated the efficacy of the proposed TriResNet in achieving increased accuracy when compared to two state-of-the-art networks.  The promising results achieved using the proposed TriResNet network indicate that such a network could be a useful tool to aid pathologists in improving the consistency, accuracy, and speed of analyzing large volumes of whole histopathology slides containing millions of cells.  In the future, we hope to leverage improved data augmentation strategies to handle some issues experienced by the proposed TriResNet network associated with staining diversity, as well as more comprehensive testing and evaluation with a larger variety of histopathology image data. Furthermore, a more comprehensive and fundamental trade-off analysis between the number of streams within the network and the level of performance achieved would be quite useful to better understand network design.

\section{Acknowledgements}
The authors thank the Natural Sciences and Engineering Research Council of Canada, the Canada Research Chairs Program, and the Queen Elizabeth II Graduate Scholarship in Science \& Technology for partially supporting this research.  Furthermore, the authors thank Nvidia for the GPUs used in this study that were provided as part of a hardware grant.

{\small
\bibliographystyle{ieee}
\bibliography{egbib}
}

\end{document}